\documentclass[runningheads]{llncs}

\usepackage{eccv}

\usepackage{eccvabbrv}

\usepackage[dvipsnames]{xcolor}
\usepackage{colortbl}

\usepackage{graphicx}
\usepackage{booktabs}

\usepackage[accsupp]{axessibility}  

\usepackage{hyperref}

\usepackage{orcidlink}
\usepackage{multirow}
\usepackage{adjustbox} 

\usepackage{pifont}
\newcommand{\cmark}{\ding{51}}
\newcommand{\xmark}{\ding{55}}
\newcommand{\greencheck}{{\color{ForestGreen}\cmark}}
\newcommand{\redcross}{{\color{BrickRed}\xmark}}
\begin{document}

\title{TimeRewind: Rewinding Time with Image-and-Events Video Diffusion} 

\titlerunning{TimeRewind}

\author{Jingxi Chen\inst{1} \and
Brandon Y. Feng\inst{2} \and
Haoming Cai\inst{1} \and  Mingyang Xie \inst{1} \and  Christopher Metzler \inst{1} \and Cornelia Ferm{\"u}ller \inst{1} \and Yiannis Aloimonos \inst{1}  }

\authorrunning{Chen et al.}

\institute{University of Maryland, College Park \and
Massachusetts Institute of Technology}

\maketitle
 
\begin{abstract}
This paper addresses the novel challenge of ``rewinding'' time from a single captured image to recover the fleeting moments missed just before the shutter button is pressed. This problem poses a significant challenge in computer vision and computational photography, as it requires predicting plausible pre-capture motion from a single static frame, an inherently ill-posed task due to the high degree of freedom in potential pixel movements. We overcome this challenge by leveraging the emerging technology of neuromorphic event cameras, which capture motion information with high temporal resolution, and integrating this data with advanced image-to-video diffusion models. 
Our proposed framework introduces an event motion adaptor conditioned on event camera data, guiding the diffusion model to generate videos that are visually coherent and physically grounded in the captured events. 
Through extensive experimentation, we demonstrate the capability of our approach to synthesize high-quality videos that effectively ``rewind'' time, showcasing the potential of combining event camera technology with generative models. Our work opens new avenues for research at the intersection of computer vision, computational photography, and generative modeling, offering a forward-thinking solution to capturing missed moments and enhancing future consumer cameras and smartphones. Please see the project page at  \textcolor[RGB]{216,16,125}{\url{https://timerewind.github.io/}} for video results and code release.
\end{abstract}

\section{Introduction}
\label{sec:intro}

\begin{figure}[t]
    \centering
    \includegraphics[width=\linewidth]{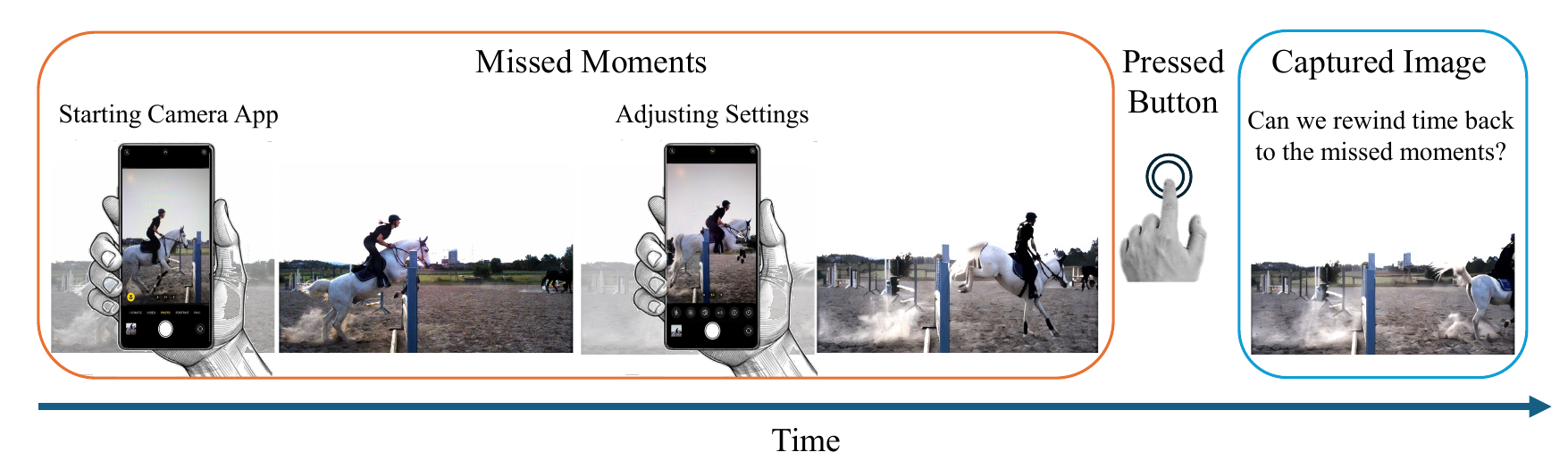}
    \caption{In the everyday use of smartphones for capturing images or videos, the process from opening the camera app to pressing the capture button involves preparation time. This includes aiming at the scene and deciding when to take the shot. Often, the moments we wish to capture occur during this preparation phase. Our work focuses on "rewinding" time from the point of capture to retrieve these missed moments.}
    \label{fig:teaser}
\end{figure}

Our daily lives are filled with countless moments that deserve to be captured and remembered. From personal highlights in sports and social events to exciting moments with our pets, each day presents unique opportunities for memorable experiences. However, most of these memorable moments are fleeting. 
As many have experienced, capturing moments with our phone's camera can often be a race against time. From the moment we decide to take a picture, launch the camera app, position our devices just right, and finally press the capture button, precious seconds tick away. By the time we press the shutter button to freeze the moments, these moments have already slipped past by. Consequently, we are often left with images that miss the mark, too late for the desired moments we were aiming for (as illustrated in Fig. \ref{fig:teaser}). This commonly experienced pain presents a challenging opportunity for computational photography: Can we turn back the clock on our captured image?

In this work, we are interested in how to rewind time from a single captured image and recover the missed moments. 
We define {\it missed moments} as the RGB video that we could have recorded, {\it after} the camera is directed towards a scene, and {\it before} the capture button is pressed. 
We need to consider three concrete questions to achieve this goal.
First, why is rewinding time from a captured image difficult?
Second, what are reasonable assumptions to make this goal practical?
Finally, what kind of computational algorithm is capable of achieving our goal?

From a single image, recovering the past motion information before capture is inherently ill-posed.
Unlike the common task of video frame {\it interpolation}, predicting a sequence of plausible frames from just a {\it single} image is far more difficult due to the excessive degrees of freedom for pixel movements.
Very recent breakthroughs in image-to-video generative models are still insufficient for our purpose.
The apparent mismatch is that video generative models are designed for forward synthesis—creating future sequences from a given frame—rather than the backward synthesis required to recapture moments that have already passed.
Furthermore, the dynamics produced by these models are not based on any physical reality, resulting in videos that may look realistic but lack grounding in actual physics.
To go beyond uncontrolled video hallucinations, we need to make more assumptions than just having access to a single image.

What reasonable assumptions can we make? To recover realistic motions, we would at least need to capture some information during that time interval.
In our situation of a hurried capture, we pull the camera out in time and may even see the scene on the screen, but we cannot press the button quickly enough.
The implication is that the light from the missed moments has reached our camera sensor, but we fail to record them.
However, this does not have to be the case in the future.
Event cameras \cite{gallego2020event} are an up-and-coming innovation for digital photography.
They only respond to motion, are power-efficient, and record at a temporal resolution of several microseconds.
We believe event cameras will be pivotal in future consumer smartphones and cameras. In this paper, we study the scenario where we have access to event camera signals during the time interval of missed moments.

\begin{figure*}[t!]
    \centering
    \includegraphics[width=.98\linewidth]{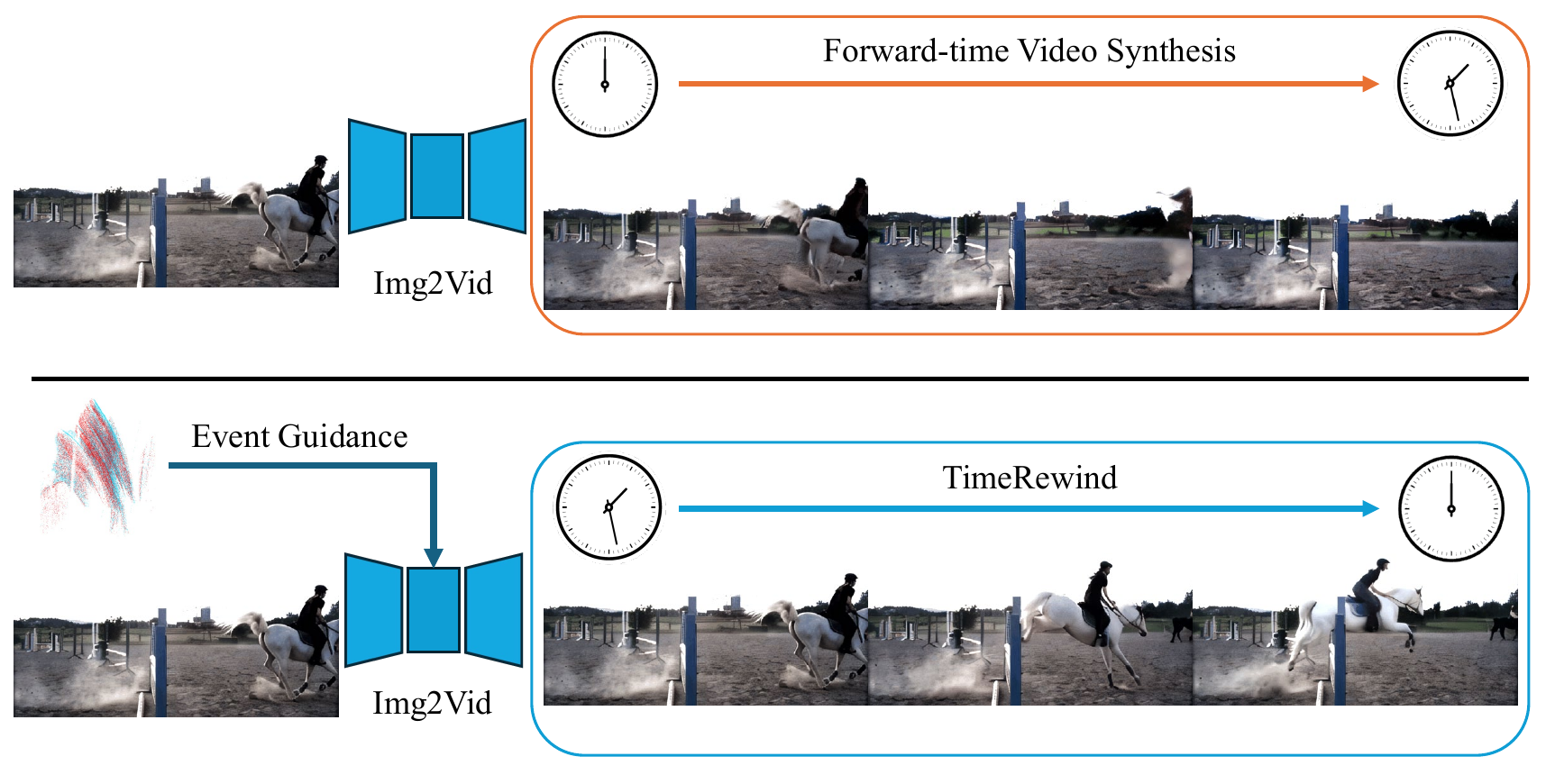}
    \caption{Img2Vid models are trained for forward-time video synthesis. In contrast, the goal of TimeRewind is to synthesize the video backward into the pre-capture time.}
    \label{fig:Intro}
\end{figure*}

What algorithm is suitable for this problem? Recent advances in video diffusion models \cite{harvey2022flexible,ho2022imagen, singer2022make, yang2023diffusion}, especially for image-to-video (Img2Vid) generation \cite{blattmann2023align, blattmann2023stable} provide an unprecedented ability to convert a single image to a video.
Img2Vid diffusion models use one image as the condition and synthesize video contents coherent with the given image condition. 
Given their impressive success, adapting the Img2Vid model framework is a promising strategy to achieve our time-rewinding goal.
However, this adaptation presents its own set of challenges. Img2Vid models are designed to use a single image as input and do not incorporate motion information (as discussed in Table \ref{tab:motion_cues}).
More importantly, they do not account for the fact that the final motion in the video is a product of camera motion, physics-based motion, and object motion.
Our proposed approach (Fig. \ref{fig:Intro}) uses events from the missed moments to guide an Img2Vid diffusion model. 
To preserve the prior knowledge in the pre-trained Img2Vid model, we keep the Img2Vid model weights frozen and only train an additional event motion adaptor. The event motion adaptor is specifically designed to fit in the diffusion process by conditioning on both events and the diffusion timestep, learning to add a residual to noise prediction at each diffusing timestep. As shown in experiments (Sec.~\ref{subsec: comparison}), our way of fusing events as additional guidance to a pre-trained Img2Vid model can achieve physics-grounded motion synthesis via events and maintain excellent Img2Vid generation quality.

In summary, this paper makes significant contributions at the intersection of computer vision, computational photography, and generative models. We pioneer the combination of emerging event camera technology with advanced video diffusion models and our contributions are listed below:
\begin{itemize}
    \item  We introduce a novel problem in computational photography: rewinding time from a single captured image to recover the missed moments before the shutter was pressed.
    \item We propose leveraging the emerging neuromorphic event camera sensor, which represents a significant departure from traditional digital cameras, and present a forward-thinking approach that anticipates the integration of event cameras into consumer smartphones and cameras.
    \item We develop a framework integrating event camera data with image-to-video diffusion models and design an event motion adaptor, ensuring the generated videos are visually coherent and physically plausible.
    \item We validate our approach through extensive experiments, showing successful time-rewind results grounded in the physics captured by event cameras while achieving high-quality generation outperforming several baseline methods.
\end{itemize}

\section{Related Work}
\subsubsection{Event Cameras.}
Event cameras are neuromorphic sensors that capture the brightness changes in the scene, and the brightness changes are primarily due to the camera's motion and object motion in the scene. For this reason, events captured by event cameras can be used for computing motion-related information, for example optical flow, motion segmentation, and ego-motion \cite{benosman2012asynchronous,barranco2014contour,bardow2016simultaneous, barranco2018real,mitrokhin2018event, mitrokhin2019ev,parameshwara20210, zhu2019unsupervised, zhu2018ev}. Various methods have been proposed to reconstruct intensity videos from event streams alone, demonstrating that while feasible, the quality of such reconstructions is significantly limited by camera and object motion \cite{cadena2023sparse, rebecq2019high}. Beyond event-only applications, combining events and images can empower unprecedented abilities for image deblurring and video frame interpolation \cite{tulyakov2021time, tulyakov2022time, zhang2022unifying, sun2023event}. These methods can achieve supervising performance on image deblurring and frame interpolation when given two images taken close in time and events in between. To the best of our knowledge, even though events are powerful and sparse representations of realistic motion, there is no work on utilizing events as motion guidance for diffusion-based video generation tasks.

\subsubsection{Diffusion Models.} 
Diffusion Models (DMs) \cite{ho2020denoising, sohl2015deep, song2020score} revolutionized generative models, particularly in generating images from textual prompts or other conditions \cite{sohl2015deep, dhariwal2021diffusion, kingma2021variational, avrahami2022blended, meng2021sdedit, brooks2023instructpix2pix, gafni2022make, hertz2022prompt, kawar2023imagic, kim2022diffusionclip, nichol2021glide, parmar2023zero, ramesh2022hierarchical, bashkirova2023masksketch, mou2023t2i, zhang2023adding, rombach2021highresolution}. DMs start from a noise sample and iteratively refine it toward a coherent target sample.
Samples produced by DMs can match given conditions, such as textual descriptions, semantic embedding, human pose, sketches, etc. Latent Diffusion Model (LDM) \cite{rombach2021highresolution} further achieves computational efficiency by encoding input samples into the latent space where latent samples are generally way smaller than the original images. 
Video diffusion models have been introduced to build upon the development and success of DMs and LDMs. They synthesize temporally consistent videos \cite{harvey2022flexible,ho2022imagen, singer2022make, yang2023diffusion}, either conditioned on text prompts or unconditioned.
Since both text-conditioned and unconditioned cases have very little control over the synthesized video contents, follow-up studies, known as  Img2Vid diffusion models \cite{blattmann2023align, blattmann2023stable}, tried to incorporate the conditional image as a stronger condition for the visual content in the synthesized videos.

\subsubsection{Motion-guided Img2Vid.} Img2Vid DMs can achieve amazing performance on visual content consistency with the conditional image, but generating motion from a single image is still an ill-posed problem with too many degrees of freedom in camera motion, object motion, and physics-based motion. As a result, an area of research focuses on how to incorporate motion guidance into Img2Vid DMs and the type of conditions used as guidance for Img2Vid DMs \cite{chen2023motion, yin2023dragnuwa, chen2023livephoto, liang2023movideo, geng2024motion, yang2024direct, zhao2023videoassembler}. Among these works, the two major types of motion guidance are text prompts \cite{chen2023livephoto, liang2023movideo, geng2024motion, yang2024direct, zhao2023videoassembler} and user-input strokes/trajectories \cite{chen2023motion,yin2023dragnuwa}. Both types are the human description of motion, which is subjective and insufficient to represent realistic motion.

\section{Method}

This section outlines our proposed methodology, TimeRewind, which leverages events as additional motion guidance for the Img2Vid DM to achieve backward-time video synthesis with realistic motion.

\begin{figure*}[t!]
    \centering
    \includegraphics[width=.98\linewidth]{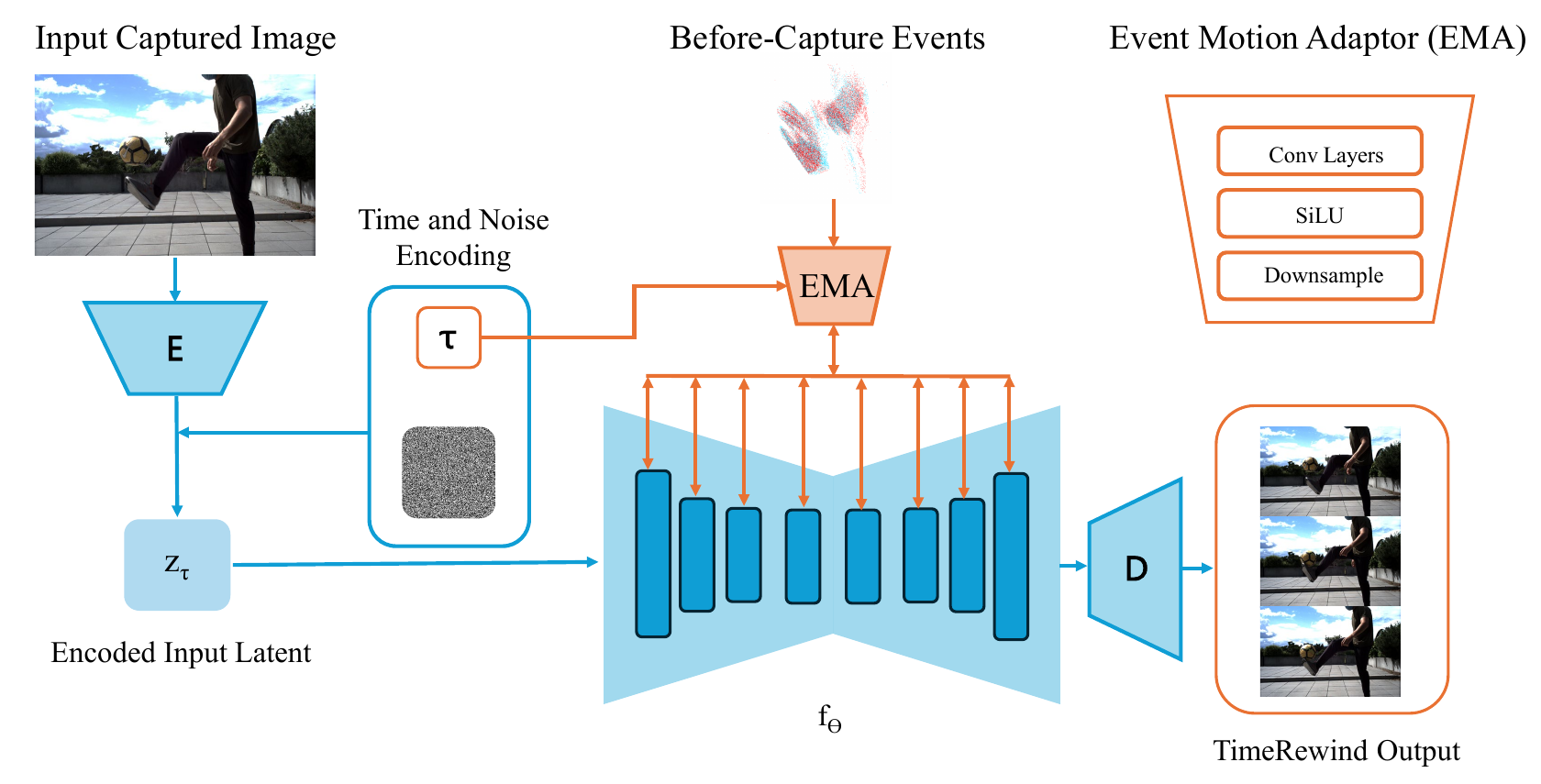}
    \caption{Illustration of our proposed TimeRewind approach as an adaptor for the general Img2Vid architectures. The components shown in shades of blue (both dark and light) represent the elements of the original pre-trained model, which remain unchanged during our training process. The orange-colored components are specific to our TimeRewind and are being optimized throughout the training.  $E$ denotes the VAE encoder to convert input captured images into latent space, $\tau$ is the diffuse time for a diffuse step, $f_{\theta}$ is the denoiser network, generally it is an UNet-like architecture with $N$ total number of up, mid and down blocks. Our EMA module contains N heads of Convolution Layers, SiLU activation \cite{hendrycks2016gaussian} and downsample layers. It takes events $e$, diffuse time $\tau$, input latent $z_{\theta}$ as input conditions. The objective of EMA is to accurately predict the residual changes necessary to transfer the motion information from the input events to the input latent $z_{\theta}$.  Through a series of iterative diffusing steps, this process ensures that the motion information is seamlessly integrated into the input latent, which is decoded into the final ``TimeRewind'' videos. }
    \label{fig:pipeline}
\end{figure*}

Our pipeline comprises two major components: a pre-trained Img2Vid DM and our trained Event Motion Adaptor (EMA) module. The EMA is specifically designed to control the pre-trained Img2vid DM models with events as the realistic motion guidance, this enables the synthesis of videos in alignment with our goal of time rewinding. The training process can be summarized as follows: The backbone pre-trained Img2Vid DM is always frozen to preserve the learned consistent video generation prior as they are learned from a substantially larger dataset of general video content compared to ours \cite{blattmann2023stable, blattmann2023align}. The learnable component in our pipeline is the EMA module, which tries to learn the residual of noise prediction for frozen backbone Img2Vid DM at each denoising step $t$ and use this residual change to copy the motion structure from events to video latent vector which finally is decoded to the desired video (Fig. \ref{fig:pipeline}).

\subsection{Img2Vid Diffusion Model}
\label{subsec:svd}

In this paper, the backbone Img2Vid DM we choose is the Stable Video Diffusion (SVD) \cite{blattmann2023stable}, because of the accessibility and performance. The backbone model can be substituted with any Img2Vid DM. As a Latent Diffusion Model (LDM) model \cite{blattmann2023align}, the original SVD's training objective can be summarized as follows:  Given diffuse time $\tau$, noise schedules: $\alpha_{\tau}$ and $\sigma_{\tau}$,  samples $x \sim  p_{data}$, the denoised input at time $\tau$ can be written as $x_{\tau}= \alpha_{\tau}x + \sigma_{\tau}\epsilon$, where $\epsilon \in N(0, I)$ is random Gaussian noise. If we denote the denoiser model in Img2Vid DM as $f_{\theta}$, parametrized by weights $\theta$, the the training objective for the original img2vid diffusion model can be written as:  

\begin{align}
   E_{x \sim p_{data}, \tau \sim p_{\tau}, \epsilon \sim N(0, I)} [\| y - f_{\theta}(z_{\tau}; c, \tau) \|^{2}_{2}],
\end{align}
where $p_{\tau}$ is a uniform distribution for sampling diffuse time $\tau$ during the training, $z_{\tau}$ is the pre-trained Variational Autoencoder (VAE) encoded latent vector of input sample $x$, $c$ is the condition for guiding the denoiser towards denoising according to the condition. For the case of  Img2Vid, $c$ is the input image condition, and the target vector $y$ is either random noise $\epsilon$ or \textit{v-prediction} \cite{salimans2022progressive} $v= \alpha_{\tau} \epsilon - \sigma_{\tau} x$. By the original formulation, the pre-trained SVD can achieve amazing coherence of video visual contents conditioned on image condition $c$. However, as illustrated in Fig. \ref{fig:Intro}, the goal of TimeRewind is to synthesize a video backward in time and achieve realistic motion rather than just visual content coherence. To achieve this  goal of backward-time video synthesis with realistic motion, we need to incorporate before-the-capture events as realistic motion guidance into the Img2Vid diffusion model.

\subsection{Event Data Representation}
\label{subsec:event_represent}

Event cameras only respond to motion. 
An event at a pixel is triggered when the logarithm of the intensity changes by a certain threshold.
The motion-only and logarithmic mechanism makes event cameras extremely power-efficient. They can achieve a temporal resolution of several microseconds and a much higher dynamic range than standard cameras.
The events are recorded and stored in an asynchronous $(x, y, p, t)$ stream. $x, y$ are spatial locations of an event on the image, $p \in \{0, 1\}$ indicates the polarity -- an increase or decrease in intensity causes an event, and $t$ indicates the timestamp of the event. 
Compared to traditional dense frame-based video, recording an asynchronous stream of events can be memory-efficient and energy-efficient \cite{gallego2020event}. These two efficiencies are also the major limitations of applications running on today's smartphones. 
A summary of comparing events as the motion cue with other types of motion cues can be found in Table \ref{tab:motion_cues}.
As we can see, events are the only sparse (efficient) motion cues that can do real capture (motion realism) and provide necessary motion information for the final image/video formation. 
While optical flow also records motion on the image plane, it cannot be captured directly, instead, it is derived from video frames, often with errors.

\begin{table}[t!]
    \centering
    \resizebox{1.0\linewidth}{!}{
    \begin{tabular}{@{}cccccccccc@{}}
    \toprule
    \multirow{2}{*}{\textbf{Motion Cues}}  \  & \multirow{2}{*}{\textbf{Density}} \ & \multirow{2}{*}{\textbf{Capture}} \  &  \multicolumn{3}{c}{\textbf{Motion Information}}  \\ 
    \cmidrule(r){4-7}
     &  &  &   Camera Motion \ & Physics-based Motion \  & Objects Motion & \\ 

        \midrule
        Text Prompts   \  & Sparse & \redcross
                & \greencheck  & \redcross & \greencheck  \\
        Depth Images  \  & Dense & \greencheck
                & \greencheck & \redcross & \greencheck  \\
        Sparse Motion Vector \cite{chen2023motion,yin2023dragnuwa, blattmann2021ipoke}  \  & Sparse & \redcross
                & \redcross & \redcross & \greencheck   \\
        Optical Flow  \  & Dense & \redcross
                & \greencheck & \greencheck  & \greencheck   \\
        \hline
        Events (ours) & Sparse  & \greencheck & \greencheck & \greencheck & \greencheck \\
        \toprule
    \end{tabular}
    }
    \captionsetup{font=small}
    \caption{Comparison of the design choice of using events as the captured motion cues before capture time with other motion cues. Density of capture, related to energy and memory efficiency (\textit{Density}). Whether this type of motion cues can be captured or not, is related to the realism of provided motion information (\textit{Capture}). What types of motion information a motion cue can provide?  (\textit{Motion Information})}
    \label{tab:motion_cues}
\end{table}

\begin{figure*}[tb]
  \centering
  \includegraphics[width=\linewidth]{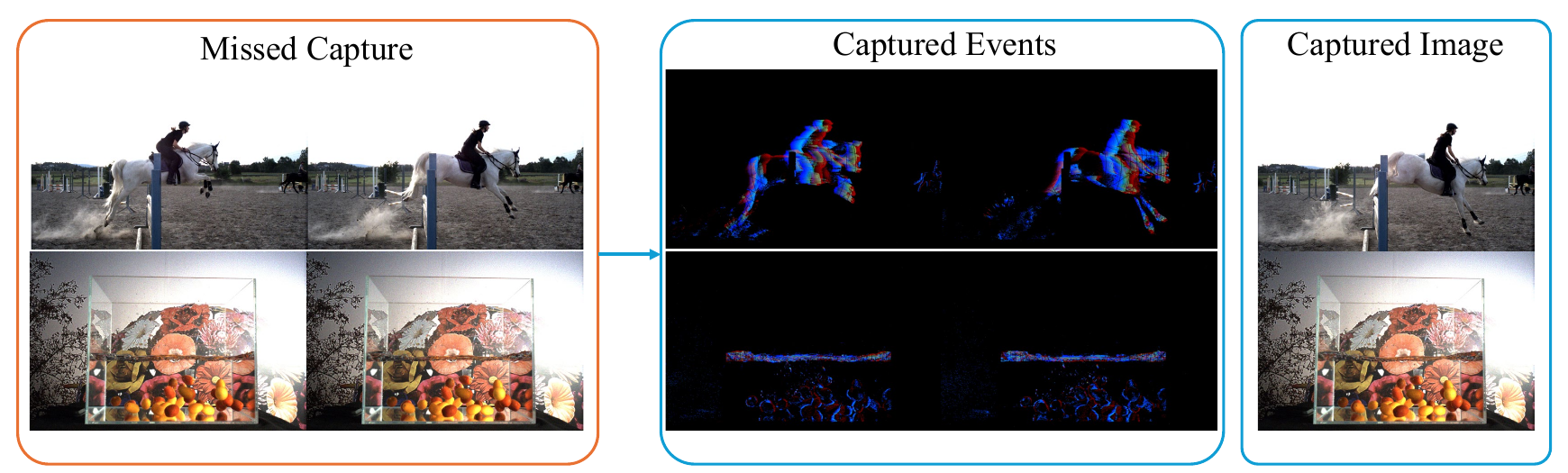}
    \caption{Illustration of our event image representation, here we show our event images accumulated from the before-capture-time events for 30 ms accumulation time window. }
    \label{fig:eventImage}
\end{figure*}

Different from frame-based data, events are recorded asynchronously as a set of $N$ events $\{ (x_{i}, y_{i}, t_{i}, p_{i} )\}_{i \in {[1, N]}}$, so we need to find a suitable representation $e$ to process and encode events. 
Inspired by event-based optical flow estimation works \cite{zhu2018ev}, we adopt the ``event image'' as our event representation $e$ by converting events during a specific accumulation time window into a colored image that encompasses both information on the number of events and timestamps of events at each pixel. The first channel of our event image is the event count image, while the second and third channels are the positive average timestamp and negative average timestamp images. An example of our event image can be seen in Fig. \ref{fig:eventImage}. For this event image representation $e$, we can reuse the pre-trained VAE encoder to convert events into the latent space for event representation $e$.  

\subsection{Event-based Motion Adaptation}
\label{subsec:ema}

As discussed in Sec.~\ref{sec:intro} and \ref{subsec:svd}, the limitations of image-only conditions and forward-time training bias in SVD prevent achieving our goal of time rewinding from a condition image. Events, capturing motion before the capture, serve as the only feasible motion guidance. Now, how can we incorporate events as motion guidance into SVD without breaking the learned video priors from its extensive training data, especially considering our much smaller dataset?  To address this, we propose freezing the pre-trained SVD model to preserve these priors and introducing an EMA module (as shown in Fig. \ref{fig:pipeline}). The EMA module adds a residual to the original SVD denoiser to copy the motion structure encoded in the event stream to the final denoised latent and decoded video. Our approach for adapting event-based motion guidance into the SVD training framework is formulated as follows:
\begin{align}
   E_{x \sim p_{data}, \tau \sim p_{\tau}, \epsilon \sim N(0, I)} [\| y - f_{\theta}(\Tilde{z}_{\phi}(z_{\tau}, g_{\phi}(e, \tau, z_{\tau})); c, \tau) \|^{2}_{2}] 
\end{align}
where latent sample vector $\Tilde{z}_{\tau}$ not only depends on the original latent sample $z_{\tau}$ but also on the learned residual latents $ g_{\phi}(e, \tau,  z_{\tau})$ from event representation $e$ and diffuse time $\tau$. To integrate event-based motion guidance within the original SVD training framework, we keep the SVD model's parameters $\theta$ unchanged during training, and we only update the parameters $\phi$ for our EMA module to convert event representation $e$ to the residual latent as the additional motion guidance for synthesizing realistic and backward-time motion in videos.

\begin{figure*}[t!]
  \centering
  \includegraphics[width=0.98\textwidth]{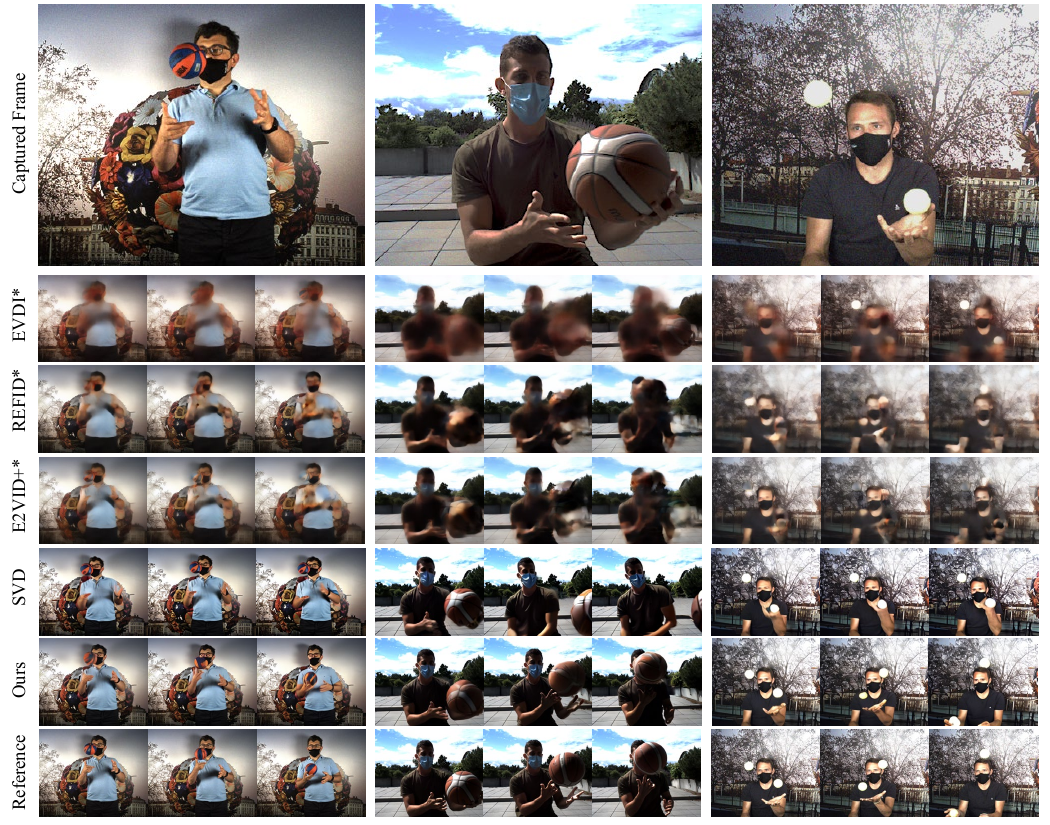}
   \caption{Comparison of backward-time video synthesis results on sequences where before-the-capture motion is simple. The motion in before-the-capture time mostly comes from a few rigid body objects. {\it Left} captures a ball being thrown upwards.  {\it Middle}  depicts a basketball spinning on a finger before being passed from the right hand to the left. {\it Right} shows juggling multiple balls between two hands.  Note that since our task is backward-time video synthesis, as in shown reference frames, the correct backward-time video sequences are the reverse process of the above descriptions.}
    \label{fig:svd_ours_1}
\end{figure*}

\section{Experiments}
\label{subsec:results}
To quantitatively and qualitatively evaluate our TimeRewind method, we adopt the standard RGB-Event dataset for training, validation, and testing, and we compare it to baseline methods.
\begin{figure*}[t!]
  \centering
  \includegraphics[width=0.98\textwidth]{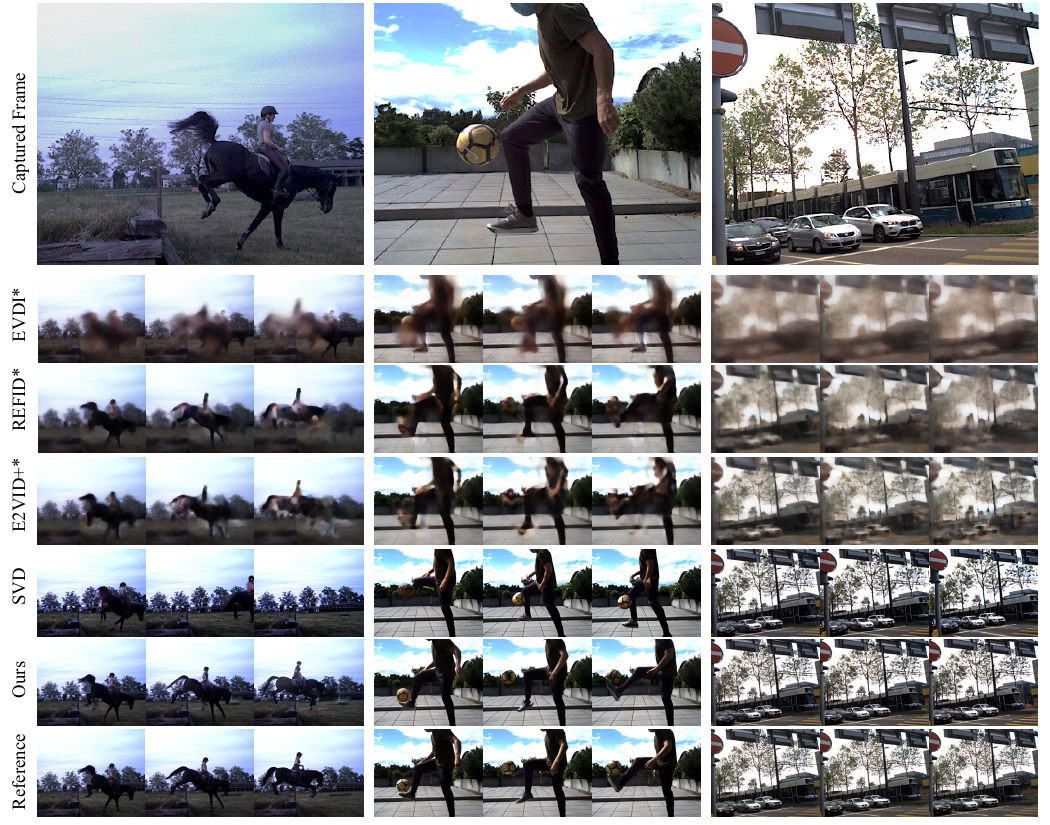}
   \caption{ Comparison of backward-time video synthesis results on sequences where before-the-capture motion is moderately complex. The motion in before-the-capture time is either due to animals and humans performing actions or scenarios with both the camera and the objects moving. {\it Left} depicts a horse descending a step. {\it Middle} shows a football being kicked upwards, skillfully caught with the foot as it descends, and then kicked up again. {\it Right}  captures a train moving forward at an intersection, with the camera vibration during filming.  Note that since our task is backward-time video synthesis, as in shown reference frames, the correct backward-time video sequences are the reverse process of the above descriptions.}
    \label{fig:svd_ours_2}
\end{figure*}

\subsection{Datasets and Training Details}
\subsubsection{Dataset.} We adopt the standard RGB-Event paired dataset BS-ERGB \cite{tulyakov2022time}. This dataset was originally used for event-based RGB video frame interpolation \cite{tulyakov2022time}. During training and testing, we only used a single frame resized to 512 $\times$ 320 (W $\times$ H) resolution and the event stream before the single frame as required by our backward-time video synthesis task.
\subsubsection{Setup.} Our baseline SVD model incorporates pre-trained weights ``stable-video-diffusion-img2vid'' from Stability AI\cite{StabilityAI2024}, designed for generating videos up to 14 frames long. To assess the effect of varying motion amounts on video generation, we adjusted the ``Motion Bucket'' hyper-parameter, integral to SVD, for modulating motion magnitude in videos. We conducted comparative analyses with the standard SVD model across a spectrum of 10 motion buckets, spaced evenly between 7 and 127, to encompass a broad range of motion in the generated videos, from minimal to substantial.

\subsubsection{Training Details.} All of our training is performed on 2 NVIDIA RTX A6000 GPUs. We train the TimeRewind model for 100,000 iterations with a learning rate of $1 \times 10^{-5}$. We train backbone network baselines with 200,000 iterations with a learning rate of $1 \times 10^{-5}$. These choices ensure the validation metrics converge and the Adam optimizer \cite{kingma2014adam} is employed.
Training typically requires around 20 hours. For inference on a single image and events data pair, the time is as follows: standard SVD with 25 denoising steps takes 13.23 seconds, TimeRewind with the same number of steps takes 15.90 seconds. In comparison, for baseline models discussed in Sec.~\ref{subsec: comparison}, EVDI* has an inference time of 1.56 seconds, REFID* takes 1.67 seconds, and E2VID+* requires 1.59 seconds.

\begin{figure*}[t!]
  \centering
  \includegraphics[width=0.98\textwidth]{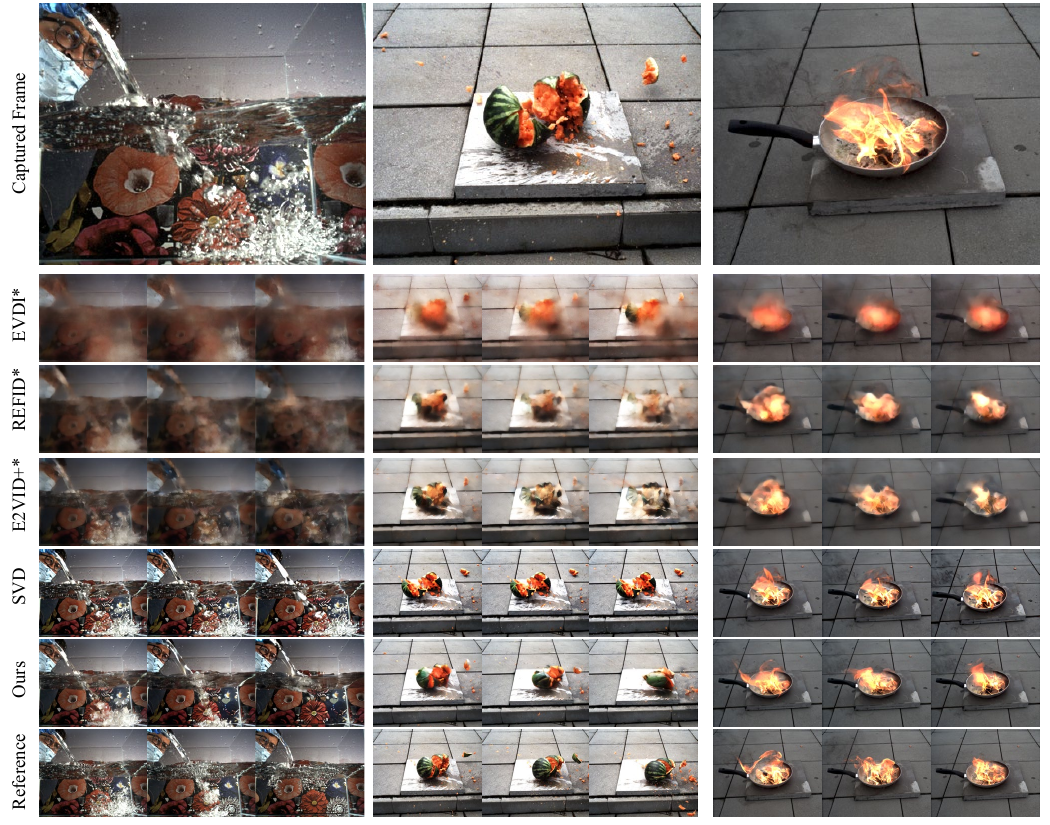}
   \caption{ Comparison of backward-time video synthesis results on sequences where before-the-capture motion is physically complex. The motion before the capture time is either due to a dynamic fluid (water stream, fire) or complex particle motion under force. {\it Left} features the formation of water streams and bubbles when pouring water into a tank. {\it Middle} shows particles dispersing from an object due to an explosion's force. {\it Right} captures a fire with flames influenced by the wind direction. Note that since our task is backward-time video synthesis, as shown in the reference frames, the correct backward-time video sequences are the reverse process of the above descriptions.}
    \label{fig:svd_ours_3}
\end{figure*}

\subsubsection{Evaluation metric.} We employ Peak Signal-to-noise Ratio (PSNR)\cite{gonzalez2009digital} and Structural Similarity Index Measure (SSIM) \cite{wang2004image} to measure the per-frame quality of synthesis. Learned Perceptual Image Patch Similarity (LPIPS) \cite{zhang2018unreasonable} is used to measure perceptual quality.

\subsection{Comparison with SVD and Choices of Backbone Architectures}
\label{subsec: comparison}
In this section, we benchmark our TimeRewind model against the standard SVD baseline and other prevalent RGB-Event multi-modal architectures, such as EVDI \cite{zhang2022unifying}, E2VID+ \cite{sun2023event} and REFID \cite{sun2023event}, which we have adapted for our unique time rewinding task by retraining them with only event data and a single image as inputs. These adapted versions are denoted as EVDI*, E2VID+*, and REFID*. Unlike their original designs, which focus on deblurring and interpolating events and RGB video frames, our model aims to rewind time. For a balanced comparison, all models utilize the SVD's VAE encoder for latent space operations, allowing us to focus on the distinct impact of each backbone architecture on the final video synthesis, independent of the VAE encoder's effects on color and perceptual quality.


\begin{table}[t!]
    \centering
    \resizebox{1.0\linewidth}{!}{
    \begin{tabular}{@{}cccccccc@{}}
    \toprule
    \multirow{2}{*}{\textbf{Method}} & \multirow{2}{*}{\textbf{Backbone}} & \multirow{2}{*}{\textbf{Finetuned}} \  &  \multicolumn{4}{c}{\textbf{Evaluation Metrics}}  \\ 
    \cmidrule(r){4-7}
     &  & & PSNR & SSIM & LPIPS \\ 
        \midrule
        SVD \cite{blattmann2023stable} & Video Diffusion \  & \redcross  & 13.93
                & 0.45   & 0.38 \\
        E2VID+* \cite{sun2023event} & Reccurent U-Net \ & \greencheck   & 20.19
                & 0.59   & 0.36 \\  
        EVDI* \cite{zhang2022unifying} & ResNet + Attention \ & \greencheck   & 18.64
                & 0.56    & 0.41  \\
        REFID* \cite{sun2023event} & Reccurent U-Net + Attention \ & \greencheck  & 20.12
                & 0.57   & 0.45  \\
        \rowcolor{pink!25}TimeRewind (Ours) & \textbf{Video Diffusion} & \greencheck   & \textbf{21.78}  & \textbf{0.70} & \textbf{0.15} \\
        \toprule
    \end{tabular}
    }
    \captionsetup{font=small}
    \caption{Comparison of our method results with the standard SVD and different choices of RGB-Event backbone architectures built upon U-Net \cite{ronneberger2015u}, ResNet \cite{he2016deep} and Attention mechanism \cite{vaswani2017attention}.}
    \label{tab:compareResults}
\end{table}

\subsubsection{Qualitative Comparison.} 
In our qualitative analysis, depicted in Fig. \ref{fig:svd_ours_1}, Fig. \ref{fig:svd_ours_2}, and Fig. \ref{fig:svd_ours_3}, we categorize motion complexity into three types.
Fig. \ref{fig:svd_ours_1} demonstrates simple motions, such as throwing, spinning, or juggling balls. Fig. \ref{fig:svd_ours_2} addresses moderate complexities, showcasing human or animal movements within a physics-based model or scenes with both camera and object movement. Here, camera movement affects the perception of object motion and triggers event camera responses on static objects. Fig. \ref{fig:svd_ours_3} explores highly complex motions influenced by fluid dynamics or force-driven particle movements, such as water flow or explosions, posing significant challenges in accurately synthesizing physics-based motion in pre-capture videos.

The visual outcomes presented in Figs. \ref{fig:svd_ours_1}, \ref{fig:svd_ours_2}, and \ref{fig:svd_ours_3} clearly demonstrate that our TimeRewind method uniquely succeeds in synthesizing pre-capture videos with motion that aligns realistically with the reference videos. Unlike the standard SVD, which inherently synthesizes forward-time videos due to biases in its training data, our method excels in backward-time video synthesis. This distinction is particularly evident in Fig. \ref{fig:svd_ours_2}, where SVD predicts the forward motion contrary to the backward movement of a horse, revealing a fundamental limitation in capturing reverse temporal dynamics. Furthermore, SVD struggles with complex motion scenarios, such as fluid and particle dynamics, because it lacks precise motion guidance. This issue becomes prominent in Fig. \ref{fig:svd_ours_3}, where SVD either avoids generating motion for fluid mediums and particles after an explosion or inaccurately shifts to camera motion to fulfill motion requirements. Conversely, our approach realistically captures the intricate physics of fluid motion in backward-time videos. For instance, in Fig. \ref{fig:svd_ours_3}, our model accurately depicts the reverse flow of water and the diminishing impact on the water in the tank, evidenced by the decreasing bubble formation, showcasing our method's superior capability to handle complex motion synthesis.

In our architecture comparison, shown in Figs. \ref{fig:svd_ours_1}, \ref{fig:svd_ours_2}, and \ref{fig:svd_ours_3}, only the standard SVD, unlike E2VID+*, EVDI*, and REFID*, avoids generating ``fog-like'' noisy and blurred videos for before-capture-time synthesis. The latter models tend to produce videos with blurred textures and pixel deformations, lacking high-level content consistency, to naively match the target videos. This highlights the limitations of these common architectures in generating content-coherent videos from an initial image. In contrast, SVD's use of a pre-trained denoiser as the backbone demonstrates its superiority due to the consistent video priors learned during its original training and its design specifically tailored for video synthesis tasks.

\subsubsection{Quantitative Comparison.} 
As shown in Table \ref{tab:compareResults}, even with different motion bucket choices, SVD fails to meet our backward-time video synthesis goal, primarily because it is designed for forward-time synthesis, leading to videos synthesized in the wrong time direction. In terms of PSNR, SSIM, and especially LPIPS metrics, our method outperforms other backbone architectures, demonstrating significantly better perceptual quality and more consistent video content with SVD as the backbone. In summary, while standard SVD's forward-time bias makes it unsuitable for our unique task, its architecture, when adapted with event data as motion guidance, enables us to synthesize consistent, backward-time videos, showcasing the advantages of leveraging pre-trained video diffusion model in our novel backward-time video synthesis task.

\section{Discussion}
\subsubsection{Practicality on consumer smartphones?}
The fundamental component of event cameras is a compact CMOS sensor that detects local brightness changes, as detailed in \cite{lichtsteiner_latency_2008,gallego2020event}. The simplicity of this sensor, combined with its superior temporal resolution and dynamic range, has prompted leading smartphone sensor manufacturers to consider integrating event sensors into future smartphone chips, as indicated by Qualcomm and Sony~\cite{Prophesee, sony}.

\subsubsection{Limitations on perceptual quality.} Since we use SVD as our backbone architecture and it is an LDM model, we encounter inherent challenges akin to those faced by similar LDMs. Specifically, when videos are encoded into the latent space and subsequently decoded back, there is an observable degradation in perceptual quality. This manifests as a loss of fine details and alterations in color fidelity.
It also highlights areas for potential improvement in maintaining the integrity of visual information through the transition between real and latent spaces for video LDMs.

\subsubsection{Limitations on missing objects/surfaces.} 
If objects are absent in the capture frame or if the visible surfaces of objects change due to rotation or more complex reasons such as an explosion, as illustrated in Fig. \ref{fig:svd_ours_3}, our TimeRewind model faces challenges. Despite having pre-captured events for motion guidance, the model may only be able to superimpose arbitrary textures that move following the motion cues. This limitation arises because motion cues alone do not provide sufficient information to accurately infer the visual appearance of previously unseen objects or surfaces.

\subsubsection{Difference from video frame interpolation.}
Event-based video frame interpolation methods \cite{tulyakov2021time, tulyakov2022time, zhang2022unifying, sun2023event} typically assume that a video sequence captured at standard frame rate already exists. Their primary objective is to utilize event data to interpolate additional frames between the existing ones, thereby enhancing the temporal resolution of the video. In contrast, TimeRewind differs significantly in its aim and application. It seeks to leverage a single image, augmented by event data, to reconstruct the moments that transpired before the shutter's activation, essentially ``rewinding'' time to generate a video sequence. Moreover, there's a notable distinction in the nature and extent of motion that our TimeRewind method can capture and reproduce. Unlike the relatively limited motion between two adjacent frames targeted by video frame interpolation techniques, TimeRewind synthesizes extended video sequences with consistent visual contents and realistic motion. 

\section{Conclusion}

In conclusion, our study introduces an innovative method to ``rewind'' time from a single image, capturing moments missed just before shutter release. Utilizing neuromorphic event cameras and image-to-video diffusion models, our framework offers a novel solution in computer vision and computational photography. It generates visually coherent and motion-accurate videos, demonstrating a significant advancement in predicting pre-capture movement. The extensive experiments conducted as part of this study have validated the efficacy of our approach, demonstrating its potential to generate high-quality videos that effectively turn back time ranging from simple to physically complex pre-capture motion scenarios. Our work validates the potential of this technology to enhance future cameras and smartphones, enabling the recapture of fleeting moments. It opens new research directions, combining event camera technology with generative models, and marks a step forward in enriching visual experiences and expanding the capabilities of consumer imaging devices.

\clearpage  

%
%
\bibliographystyle{splncs04}
\bibliography{main}
\end{document}